# Automated segmentation of an intensity calibration phantom in clinical CT images using a convolutional neural network

Keisuke Uemura[1, 2*], Yoshito Otake[1], Masaki Takao[3], Mazen Soufi[1], Akihiro Kawasaki[1], Nobuhiko Sugano[2], Yoshinobu Sato[1]

[1] Division of Information Science, Graduate School of Science and Technology, Nara Institute of Science and Technology, Ikoma city, Nara, Japan

[2] Department of Orthopaedic Medical Engineering, Osaka University Graduate School of Medicine, Suita city, Osaka, Japan

[3] Department of Orthopaedics, Osaka University Graduate School of Medicine, Suita city, Osaka, Japan

**\*Corresponding author**

Keisuke Uemura, MD, PhD

**Acknowledgments**

This study was supported by the Japan Society for the Promotion of Science Grants-in-Aid for Scientific Research (KAKENHI) Numbers 19H01176 and 20H04550. The authors would like to thank Tatsuya Kitaura MD for his help in the acquisition of data.

**Abstract**

**Purpose**

To apply a convolutional neural network (CNN) to develop a system that segments intensity calibration phantom regions in computed tomography (CT) images, and to test the system in a large cohort to evaluate its robustness.

**Methods**

A total of 1040 cases (520 cases each from two institutions), in which an intensity calibration phantom (B-MAS200, Kyoto Kagaku, Kyoto, Japan) was used, were included herein. A training dataset was created by manually segmenting the regions of the phantom for 40 cases (20 cases each). Segmentation accuracy of the CNN model was assessed with the Dice coefficient and the average symmetric surface distance (ASD) through the 4-fold cross validation. Further, absolute differences of radiodensity values (in Hounsfield units: HU) were compared between manually segmented regions and automatically segmented regions. The system was tested on the remaining 1000 cases. For each institution, linear regression was applied to calculate coefficients for the correlation between radiodensity and the densities of the phantom.

**Results**

After training, the median Dice coefficient was 0.977, and the median ASD was 0.116 mm. When segmented regions were compared between manual segmentation and automated segmentation, the median absolute difference was 0.114 HU. For the test cases, the median correlation coefficient was 0.9998 for one institution and was 0.9999 for the other, with a minimum value of 0.9863.

**Conclusions**

The CNN model successfully segmented the calibration phantom's regions in the CT images with excellent accuracy, and the automated method was found to be at least equivalent to the conventional manual method. Future study should integrate the system by automatically segmenting the region of interest in bones such that the bone mineral density can be fully automatically quantified from CT images. The source code and the model used for segmenting the phantom are open and can be accessed via https://github.com/keisuke-uemura/CT-Intensity-Calibration-Phantom-Segmentation.

**Introduction**

Quantification of bone mineral density (BMD) is necessary in the diagnosis of osteopenia and osteoporosis. Usually, lumbar vertebral or proximal femoral BMD is quantified using dual-energy X-ray absorptiometry—the procedure recommended by the World Health Organization and by several guidelines [1-3]. Yet, BMD assessment in other specific regions is also important as it can be used for surgical planning to achieve good clinical results; to this end, studies have used quantitative computed tomography (CT) images to determine local BMD in the proximal femur [4], femoral head [5,6], and the distal radius [7].

To quantify BMD using CT, an intensity calibration phantom that contains known densities of either hydroxyapatite, $Ca_{10}(PO_4)_6(OH)_2$, or dipotassium hydrogenphosphate, $K_2HPO_4$, must be included in the field of view to be able to convert radiodensity, in Hounsfield units (HU), to bone density, in $mg/cm^3$. This conversion is necessary when comparing CT results between patients and between institutions because studies have shown that the type of CT device, imaging protocol (e.g., tube voltage and slice thickness), and reconstruction protocol (e.g., convolution kernel) affect HU values [8-10]. Conventionally, researchers manually select calibration phantom regions of interest on CT images, measure radiodensity within each region of interest, and apply a linear regression model to convert radiodensity values into tissue density values [4-6,10]; however, this process is time consuming and prone to intra- and inter-operator variability, which should be avoided in multicenter studies with large datasets.

In this study, we aimed 1) to develop a system that automatically segments the calibration phantom's regions of interest (in the CT image) and converts HU into $mg/cm^3$, and 2) to evaluate the accuracy and robustness of the system by using CT images acquired at different institutions.

**Materials and methods**

A total of 1040 cases, data from patients who underwent hip surgery at two institutions (n=520 each, denoted herein as hospitals A and B) were included in this retrospective study. Ethics approval was obtained from the institutional review boards of each institutes that participated in this study. The primary reasons for hip surgery were, in hospital A, osteoarthritis (n=390), osteonecrosis (n=69), and implant loosening (n=26), and, in hospital B, proximal femoral fracture (n=511). At both hospitals, preoperative CT images are routinely acquired with an intensity calibration phantom (B-MAS200, Kyoto Kagaku, Kyoto, Japan) placed under the patient's body, approximately under the hip (Fig. 1a). This phantom is made from urethane foam (0 $mg/cm^3$) and contains four hydroxyapatite rods with known densities (50 $mg/cm^3$, 100 $mg/cm^3$, 150 $mg/cm^3$, and 200 $mg/cm^3$). The manufacturer and model of the CT device and the imaging protocols used in hospitals A and B are shown in Table 1. CT image matrix and voxel sizes were similar between the hospitals.

*Segmentation of the calibration phantom*

Because of the physical flexibility of the materials of which the calibration phantom is composed, the phantom deforms under the patient's weight (Fig. 1b). Thus, though the same model of calibration phantom was used throughout the study, segmentation of the CT image using a simple rigid 3D phantom model definition was not possible. Instead, we employed Bayesian U-Net [11], a convolutional neural network for semantic segmentation. For the training dataset, we randomly selected 40 cases (20 cases from each hospital), and in each image, the calibration phantom was manually segmented using Synapse Vincent software (v4.4, Fujifilm, Tokyo, Japan).

*Automated calibration and post-processing*

Analysis of the convolution neural network's segmentation results was performed on the remaining 1000 cases (n=500 from each hospital). After the regions calibration phantom were defined (Fig. 2a), the segmented regions were eroded using a 3-pixel disk-shaped structuring element to avoid susceptibility to small variations at the boundaries that can affect the values of radiodensity measured in each material (Fig. 2b). Linear regression (the standard protocol suggested by the manufacturer) was applied to model the relationships between HU and $mg/cm^3$ for each test case (Fig. 2c); the slopes and correlation coefficients of the regression models were calculated. Matlab (v9.8, The MathWorks, Natick, MA, USA) was used for post-processing and the calibration process.

*Quantitative assessment of segmentation accuracy*

To assess segmentation accuracy, 4-fold cross validation was performed on the training dataset—15 cases from each hospital (n=30 total) were randomly selected for training and the remaining five cases from each hospital (n=10 total) were used for validation in each fold. Accuracy was evaluated using the Dice coefficient [12] and average symmetric surface distance (ASD) [13]. Furthermore, to determine the effect of segmentation method differences, absolute differences of radiodensity values were compared between manual segmentation and automated segmentation regions.

### *Calculation of density using the regression models*

To analyze the effect of calibration, the linear regression models were applied to the range between −100 and 700 HU (the radiodensity range of human tissues often used as the target of clinical analysis), and the results were compared between hospitals.

### *Comparison between the conventional manual method and the automated method*

The regression models of the automated method were compared with regression models with the conventional manual method. Specifically, circular-shaped regions of interest were manually defined for each rod on three axial CT slices, and a linear regression model was generated using the mean values for the three slices (Fig. 3). Correlation coefficients were compared between the conventional manual method and the automated method.

### *Statistical analysis*

Normality was assessed with a Shapiro-Wilk test; data were expressed as mean ± standard deviation when normally distributed and as median (interquartile range) when not normally distributed. Data were compared using the Mann-Whitney U-test when not normally distributed. The Benjamini-Hochberg procedure was used to correct for multiple comparisons. To compare paired non-normally distributed data, the Wilcoxon signed-rank test was used. All statistical analyses were performed using Matlab and values of $p<0.05$ were considered statistically significant.

**Results**

*Segmentation accuracy*

Automated segmentation was possible in all 1000 test cases including challenging cases with severe artifacts as a result of metal implants in the field of view and cases with the phantom partially located outside of the field of view (Fig. 4). The source code and the model used for segmenting the phantom are open and can be accessed via https://github.com/keisuke-uemura/CT-Intensity-Calibration-Phantom-Segmentation.

After 4-fold cross validation, the median Dice coefficient was 0.977 (0.023), and the median ASD was 0.116 mm (0.108) (Fig. 5). No significant difference between hospitals was found for the Dice coefficient (p=0.84), but the ASD for hospital A was significantly larger than that for hospital B (p=0.02) (Table 2). When segmented regions were compared between manual segmentation and automated segmentation, the median absolute difference was 0.142 HU (0.280) for hospital A and 0.082 HU (0.124) for hospital B, an overall absolute difference of 0.114 HU (0.158) (Table 2).

*Comparison between hospital regression models*

The median correlation coefficient of the regression was 0.9998 (0.0004) for hospital A and 0.9999 (0.0001) for hospital B. The median slope of the regression model was 0.841 (0.027) for hospital A and 0.744 (0.041) for hospital B (Fig. 6), which was significantly different (p<0.001). For the range between −100 and 700 HU, there were significant differences between the hospitals for the ranges between −100 and 4 HU and between 11 and 700 HU—the difference in the result for tissue density was 0.6 mg/cm$^3$ for 0 HU, which expanded to 58.2 mg/cm$^3$ for 600 HU (Fig. 4).

***Comparison between the conventional manual method and the automated method***

The median correlation coefficient of the regression models was 0.9999 (0.0002) for the conventional manual method and 0.9999 (0.0001) for the automated method. No significant difference was found between the methods (p=0.12).

**Discussion**

We applied a convolutional neural network to automatically segment the corresponding regions of differing radiodensity that corresponded to the different known tissue densities of an intensity calibration phantom used in clinical CT images. The model accuracy—Dice coefficient (0.977) and ASD (0.116 mm)—and robustness—the overall absolute difference between manual and automated segmentations (0.114 HU)—were excellent after training; however, when tissue densities were calculated, significant differences were found between the two hospitals, especially for larger values in the typical clinical range.

In quantitative CT analyses, researchers typically select the calibration phantom manually from a few axial slices of the CT images to create a regression model between radiodensity and tissue density. When the correlation coefficients of the regression models were compared between the conventional manual method and the automated method, no significant difference was found. Importantly, the median correlation coefficients of both methods were higher than 0.999. Thus, the automated method is at least equivalent to the conventional manual method for developing calibration regression models. Since the automated method does not require a manual process (selecting axial slices, selecting regions of interest, and creating regression models), the automated system saves time and suits for multicenter studies where many researchers participate.

There were significant differences between the tissue densities calculated with hospital A's regression model and hospital B's regression model when CT image radiodensity ranged from −100 to 4 and from 11 to 700. Small differences, even if statistically significant, may not be clinically important. However, because of the difference in the slope of the regression models (Fig. 4), the differences in tissue density value increased as the radiodensity value increased—resulting in a difference of 58.2 mg/cm$^3$ at 600 HU (Fig. 4). This finding is in line with Giambini et al's finding [9], an important finding that large errors are expected if the static range definition of 300–600 HU is used to define cortical bone in CT images, as has been done in previous bone surface modeling and finite element studies [14-17]. We suggest that static definitions should not be used between institutes and recommend using an intensity calibration phantom when comparing BMD values between institutes.

Recently, studies have employed convolutional neural networks to diagnose hip diseases [18,19] and to segment musculoskeletal regions in CT images [11]. To the best of our knowledge, no studies have developed an automated system that uses a convolutional neural network to segment the regions of imaging calibration phantoms in CT images, which makes our study novel and clinically important. In recent studies, a phantom-less calibration method using internal reference tissues of each patient, such as the aortic blood tissue, pelvic visceral adipose tissue, muscle, and fat, have been reported [10,20] with conflicting results—one paper reported the usefulness of the method [10] while the other recommended caution [20]. Importantly, these previous studies included only a limited number of cases because manual selection of the reference and of the regions of interest was necessary. It would be interesting to apply the system developed in this study to clarify the usefulness/accuracy of the phantom-less calibration method.

It is important to note that manual efforts are still necessary to quantify BMD from CT images because the bone regions of interest must be selected manually. Because this process is time consuming and prone to error [21], in future studies, we aim to explore developing a convolutional neural network to isolate the region of interest (e.g., femoral neck and spine) to create a fully automated system, which would pave the way for conducting multicenter quantitative CT studies with large datasets.

### *Limitations*

There were some limitations in this study. First, while the system was tested at two hospitals with different CT devices, imaging protocols, and reconstruction protocols including two types of convolution kernels (i.e., soft tissue and bone) that are commonly used in the field of orthopedics, results may vary if CT images are acquired in different situations. However, it is likely that, if an appropriate training dataset is added, the system would be able to perform sufficiently. Second, because only axial slices were used, images with halation (e.g., metal artifacts and beam hardening) were included (Fig. 6); however, the effect of halation on the regression models was weakened (because there were relatively few of these images) and was likely negligible; this assumption is supported by the high minimum correlation coefficient (0.9863).

## Conclusions

The convolutional neural network was able to accurately segment the intensity calibration phantom from the CT images with a mean Dice coefficient of 0.977, ASD of 0.116 mm, and an error of 0.114 HU. The median correlation coefficient of the regression models was greater than 0.999, which demonstrated the excellent ability of the developed system to convert radiodensity into tissue density. Large differences between models were found for tissue densities based on inputs in the range defined for cortical bone (300–600 HU), indicating the necessity of using a calibration phantom to compare the results between institutions. Our next step will be to integrate this system with a system for the automated segmentation of region of interest in bones [11] to quantify BMD, for a fully automatic system.

**Declarations:**

**Funding**

This study was supported by the Japan Society for the Promotion of Science (JSPS) Grants-in-Aid for Scientific Research (KAKENHI) Numbers 19H01176 and 20H04550.

**Competing interests**

The authors have nothing to disclose.

**Ethics approval**

All procedures performed in this study was performed in accordance with the ethical standards as laid down in the 1964 Declaration of Helsinki and its later amendments or comparable ethical standards.

**Informed consent**

This study was approved by the institutional review board, and written informed consent was waived because of the retrospective design.

**Availability of data and material**

Model used for phantom segmentation can be accessed via https://github.com/keisuke-uemura/CT-Intensity-Calibration-Phantom-Segmentation

**Code availability**

Code used for phantom segmentation can be accessed via https://github.com/keisuke-uemura/CT-Intensity-Calibration-Phantom-Segmentation

**Authors' contributions**

Conceptualization: Keisuke Uemura, Yoshito Otake; Methodology: Keisuke Uemura, Yoshito Otake; Code writing: Mazen Soufi; Formal analysis and investigation: Keisuke Uemura, Akihiro Kawasaki; Writing - original draft preparation: Keisuke Uemura; Writing - review and editing: Yoshito Otake, Masaki Takao, Mazen Soufi, Nobuhiko Sugano, Yoshinobu Sato; Funding acquisition: Yoshito Otake, Yoshinbu Sato. All authors read and approved the final manuscript.

**References**

1. Kanis JA, Cooper C, Rizzoli R, Reginster JY (2019) European guidance for the diagnosis and management of osteoporosis in postmenopausal women. Osteoporosis international : a journal established as result of cooperation between the European Foundation for Osteoporosis and the National Osteoporosis Foundation of the USA 30 (1):3-44. doi:10.1007/s00198-018-4704-5

2. Orimo H, Nakamura T, Hosoi T, Iki M, Uenishi K, Endo N, Ohta H, Shiraki M, Sugimoto T, Suzuki T, Soen S, Nishizawa Y, Hagino H, Fukunaga M, Fujiwara S (2012) Japanese 2011 guidelines for prevention and treatment of osteoporosis--executive summary. Archives of osteoporosis 7 (1-2):3-20. doi:10.1007/s11657-012-0109-9

3. Camacho PM, Petak SM, Binkley N, Diab DL, Eldeiry LS, Farooki A, Harris ST, Hurley DL, Kelly J, Lewiecki EM, Pessah-Pollack R, McClung M, Wimalawansa SJ, Watts NB (2020) AMERICAN ASSOCIATION OF CLINICAL ENDOCRINOLOGISTS/AMERICAN COLLEGE OF ENDOCRINOLOGY CLINICAL PRACTICE GUIDELINES FOR THE DIAGNOSIS AND TREATMENT OF POSTMENOPAUSAL OSTEOPOROSIS-2020 UPDATE. Endocrine practice : official journal of the American College of Endocrinology and the American Association of Clinical Endocrinologists 26 (Suppl 1):1-46. doi:10.4158/gl-2020-0524suppl

4. Maeda Y, Sugano N, Saito M, Yonenobu K (2011) Comparison of femoral morphology and bone mineral density between femoral neck fractures and trochanteric fractures.

Clinical orthopaedics and related research 469 (3):884-889. doi:10.1007/s11999-010-1529-8

5. Uemura K, Takao M, Otake Y, Hamada H, Sakai T, Sato Y, Sugano N (2018) The distribution of bone mineral density in the femoral heads of unstable intertrochanteric fractures. Journal of orthopaedic surgery (Hong Kong) 26 (2):2309499018778325. doi:10.1177/2309499018778325

6. Whitmarsh T, Otake Y, Uemura K, Takao M, Sugano N, Sato Y (2019) A cross-sectional study on the age-related cortical and trabecular bone changes at the femoral head in elderly female hip fracture patients. Scientific reports 9 (1):305. doi:10.1038/s41598-018-36299-y

7. Hanusch BC, Tuck SP, Mekkayil B, Shawgi M, McNally RJQ, Walker J, Francis RM, Datta HK (2020) Quantitative Computed Tomography (QCT) of the Distal Forearm in Men Using a Spiral Whole-Body CT Scanner - Description of a Method and Reliability Assessment of the QCT Pro Software. Journal of clinical densitometry : the official journal of the International Society for Clinical Densitometry 23 (3):418-425. doi:10.1016/j.jocd.2019.05.005

8. Adams JE (2009) Quantitative computed tomography. European journal of radiology 71 (3):415-424. doi:10.1016/j.ejrad.2009.04.074

9. Giambini H, Dragomir-Daescu D, Huddleston PM, Camp JJ, An KN, Nassr A (2015) The Effect of Quantitative Computed Tomography Acquisition Protocols on Bone Mineral Density Estimation. Journal of biomechanical engineering 137 (11):114502.

doi:10.1115/1.4031572

10. Lee DC, Hoffmann PF, Kopperdahl DL, Keaveny TM (2017) Phantomless calibration of CT scans for measurement of BMD and bone strength-Inter-operator reanalysis precision. Bone 103:325-333. doi:10.1016/j.bone.2017.07.029

11. Hiasa Y, Otake Y, Takao M, Ogawa T, Sugano N, Sato Y (2020) Automated Muscle Segmentation from Clinical CT Using Bayesian U-Net for Personalized Musculoskeletal Modeling. IEEE transactions on medical imaging 39 (4):1030-1040. doi:10.1109/tmi.2019.2940555

12. Dice LR (1945) Measures of the Amount of Ecologic Association Between Species. Ecology 26 (3):297-302. doi:10.2307/1932409

13. Styner M, Lee J, Chin B, Chin M, Commowick O, Tran H, Markovic-Plese S, Jewells V, Warfield S (2008) 3D Segmentation in the Clinic: A Grand Challenge II: MS lesion segmentation. Midas Journal:1-5

14. Aamodt A, Kvistad KA, Andersen E, Lund-Larsen J, Eine J, Benum P, Husby OS (1999) Determination of Hounsfield value for CT-based design of custom femoral stems. The Journal of bone and joint surgery British volume 81 (1):143-147

15. Gausden EB, Nwachukwu BU, Schreiber JJ, Lorich DG, Lane JM (2017) Opportunistic Use of CT Imaging for Osteoporosis Screening and Bone Density Assessment: A Qualitative Systematic Review. The Journal of bone and joint surgery American volume 99 (18):1580-1590. doi:10.2106/jbjs.16.00749

16. Kitamura K, Fujii M, Utsunomiya T, Iwamoto M, Ikemura S, Hamai S, Motomura G,

Todo M, Nakashima Y (2020) Effect of sagittal pelvic tilt on joint stress distribution in hip dysplasia: A finite element analysis. Clinical biomechanics (Bristol, Avon) 74:34-41. doi:10.1016/j.clinbiomech.2020.02.011

17. Schreiber JJ, Anderson PA, Rosas HG, Buchholz AL, Au AG (2011) Hounsfield units for assessing bone mineral density and strength: a tool for osteoporosis management. The Journal of bone and joint surgery American volume 93 (11):1057-1063. doi:10.2106/jbjs.j.00160

18. Mawatari T, Hayashida Y, Katsuragawa S, Yoshimatsu Y, Hamamura T, Anai K, Ueno M, Yamaga S, Ueda I, Terasawa T, Fujisaki A, Chihara C, Miyagi T, Aoki T, Korogi Y (2020) The effect of deep convolutional neural networks on radiologists' performance in the detection of hip fractures on digital pelvic radiographs. European journal of radiology 130:109188. doi:10.1016/j.ejrad.2020.109188

19. Cheng CT, Ho TY, Lee TY, Chang CC, Chou CC, Chen CC, Chung IF, Liao CH (2019) Application of a deep learning algorithm for detection and visualization of hip fractures on plain pelvic radiographs. European radiology 29 (10):5469-5477. doi:10.1007/s00330-019-06167-y

20. Therkildsen J, Thygesen J, Winther S, Svensson M, Hauge EM, Böttcher M, Ivarsen P, Jørgensen HS (2018) Vertebral Bone Mineral Density Measured by Quantitative Computed Tomography With and Without a Calibration Phantom: A Comparison Between 2 Different Software Solutions. Journal of clinical densitometry : the official journal of the International Society for Clinical Densitometry 21 (3):367-374.

doi:10.1016/j.jocd.2017.12.003

21. Feit A, Levin N, McNamara EA, Sinha P, Whittaker LG, Malabanan AO, Rosen HN (2019) Effect of Positioning of the Region of Interest on Bone Density of the Hip. Journal of clinical densitometry : the official journal of the International Society for Clinical Densitometry. doi:10.1016/j.jocd.2019.04.002

**Table 1** CT equipment, imaging protocol, and image characteristics of hospitals A and B

| Hospital | CT manufacturer (model) | Tube voltage (kVp) | Convolution kernel (type) | Matrix size | Voxel size (mm) |
|---|---|---|---|---|---|
| A | General Electric (Optima CT660) | 120 | Standard (soft tissue) | 512 × 512 | (0.703-0.977) × (0.703-0.977) × (1.0-2.5) |
| B | Toshiba (Activion16) | | FC30 (bone) | | (0.622-0.972) × (0.622-0.972) × (0.5-2.0) |

**Table 2** Results of the 4-fold cross validation

| Parameter | Overall | Hospital A | Hospital B | p value |
|---|---|---|---|---|
| Dice coefficient | 0.977 (0.023) | 0.977 (0.025) | 0.977 (0.017) | 0.84 |
| ASD (mm) | 0.116 (0.108) | 0.136 (0.208) | 0.106 (0.073) | 0.02 |
| Absolute difference in HU | 0.114 (0.158) | 0.142 (0.280) | 0.082 (0.124) | 0.003 |

Data are expressed as median (interquartile range).

ASD: average symmetric surface distance, HU: Hounsfield units

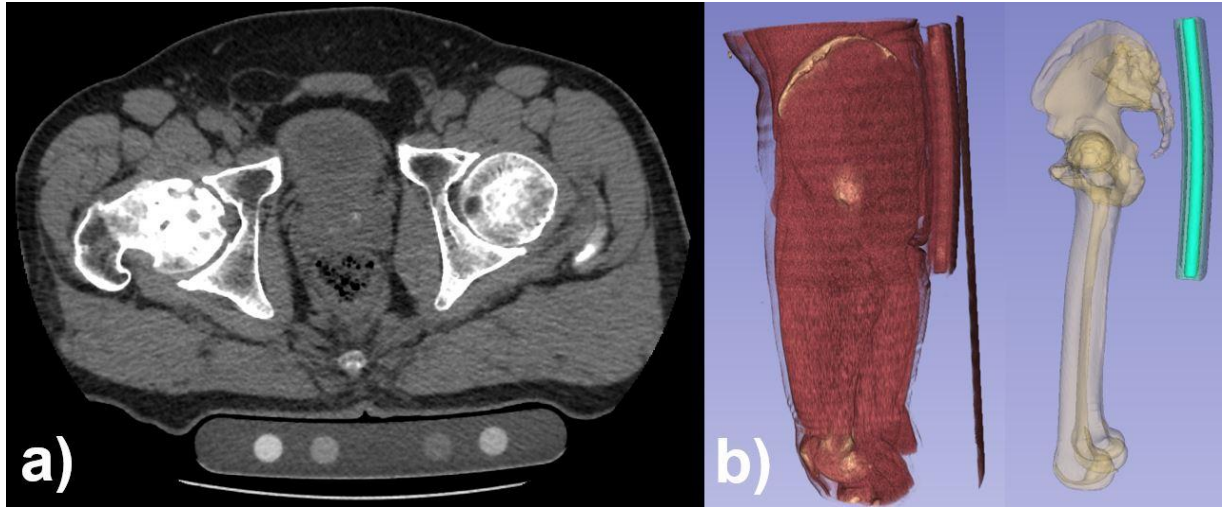

**Fig. 1** An intensity calibration phantom, placed under the hip, that was included (a) in the field of view of an axial CT image at the level of the center of the femoral head and (b) in a lateral view of volume rendering of the CT images (left) and segmented bones (right) shows deformation as a result of the patient's weight

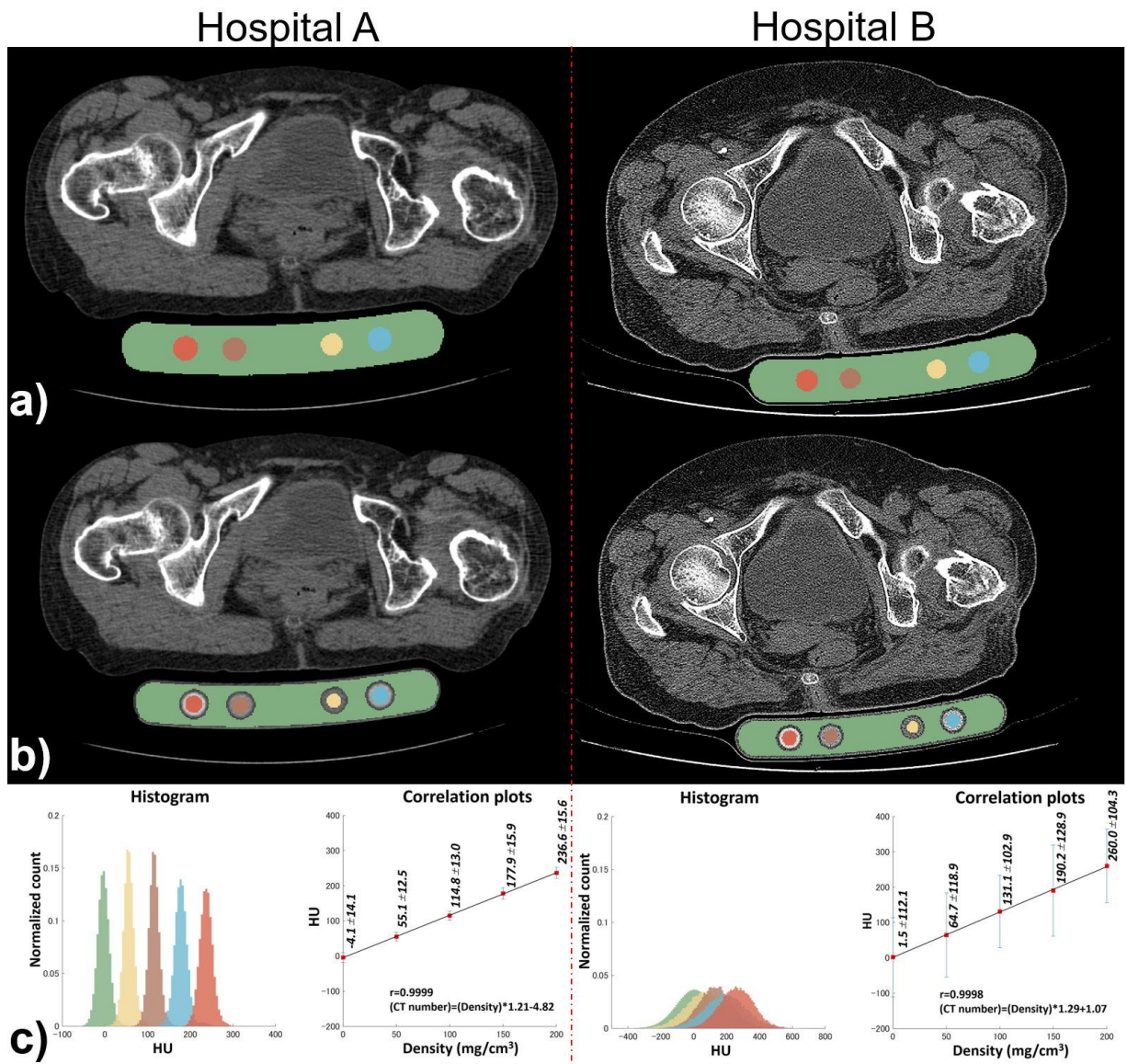

**Fig. 2** Example CT images from hospital A (left) and hospital B (right) show the intensity calibration phantom was (a) segmented into regions (representing 0 mg/cm$^3$, 50 mg/cm$^3$, 100 mg/cm$^3$, 150 mg/cm$^3$, and 200 mg/cm$^3$ indicated by green, yellow, brown, cyan, and vermilion, respectively) using Bayesian U-Net and (b) the regions in the image were filled and eroded. Example plots show (c) histograms of each region's radiodensity were calculated (left); and a linear regression model was applied, correlation coefficients were calculated, and the equation was determined (right)

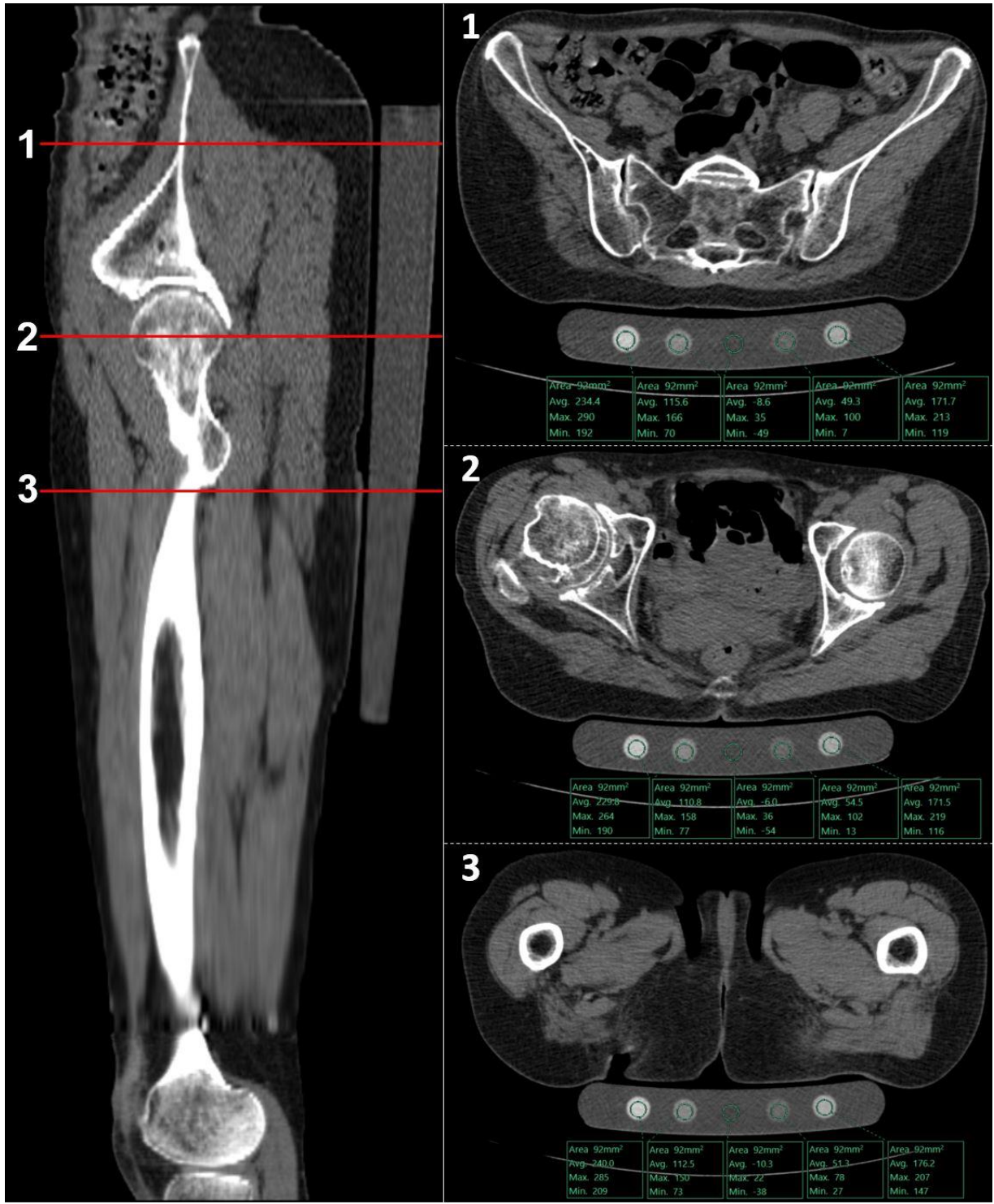

**Fig. 3** Manual measurement of calibration phantom radiodensity was performed on three axial slices (right) indicated by red horizontal lines labeled 1, 2, and 3 on the sagittal view (left)

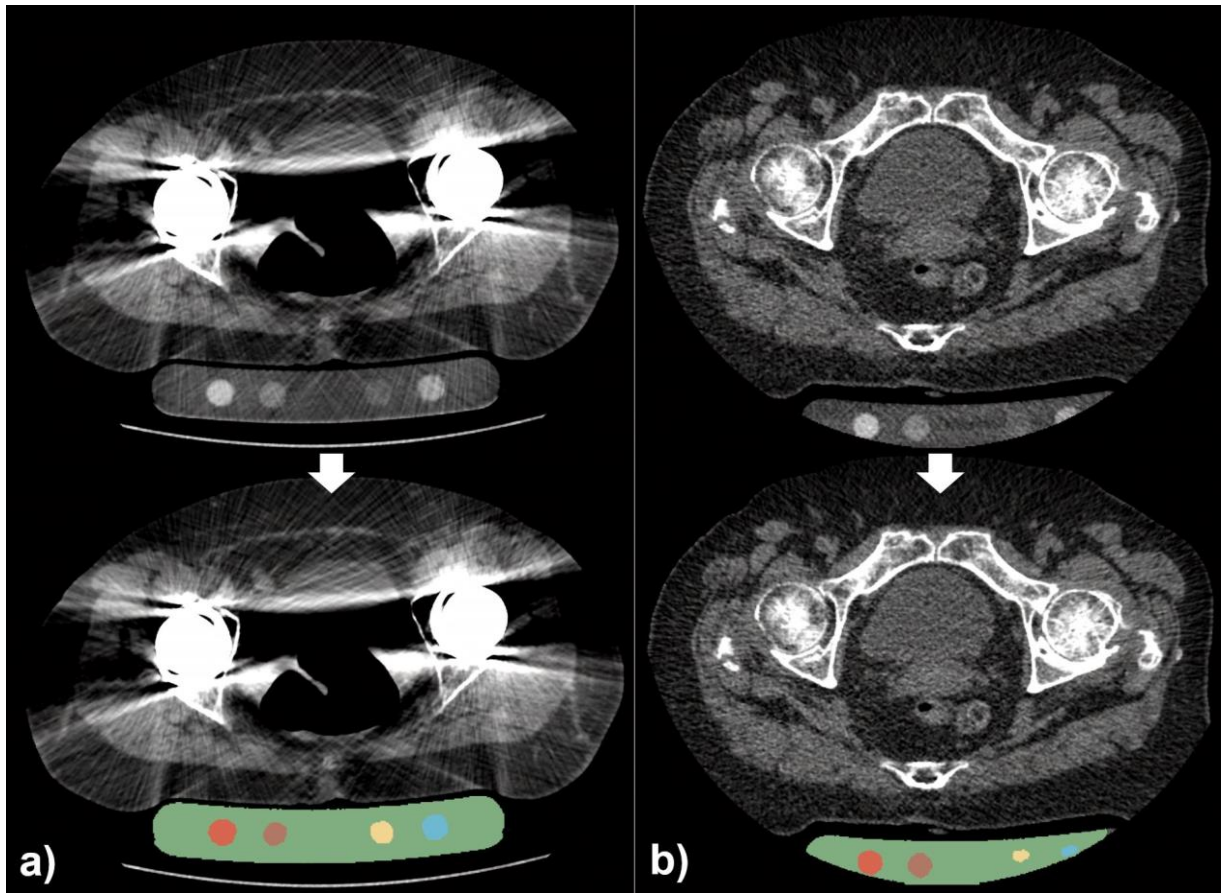

**Fig. 4** Two examples of the challenging cases: (a) image artifacts because of bilateral metallic hip implants and (b) an image in which the calibration phantom is partially located outside of the field of view

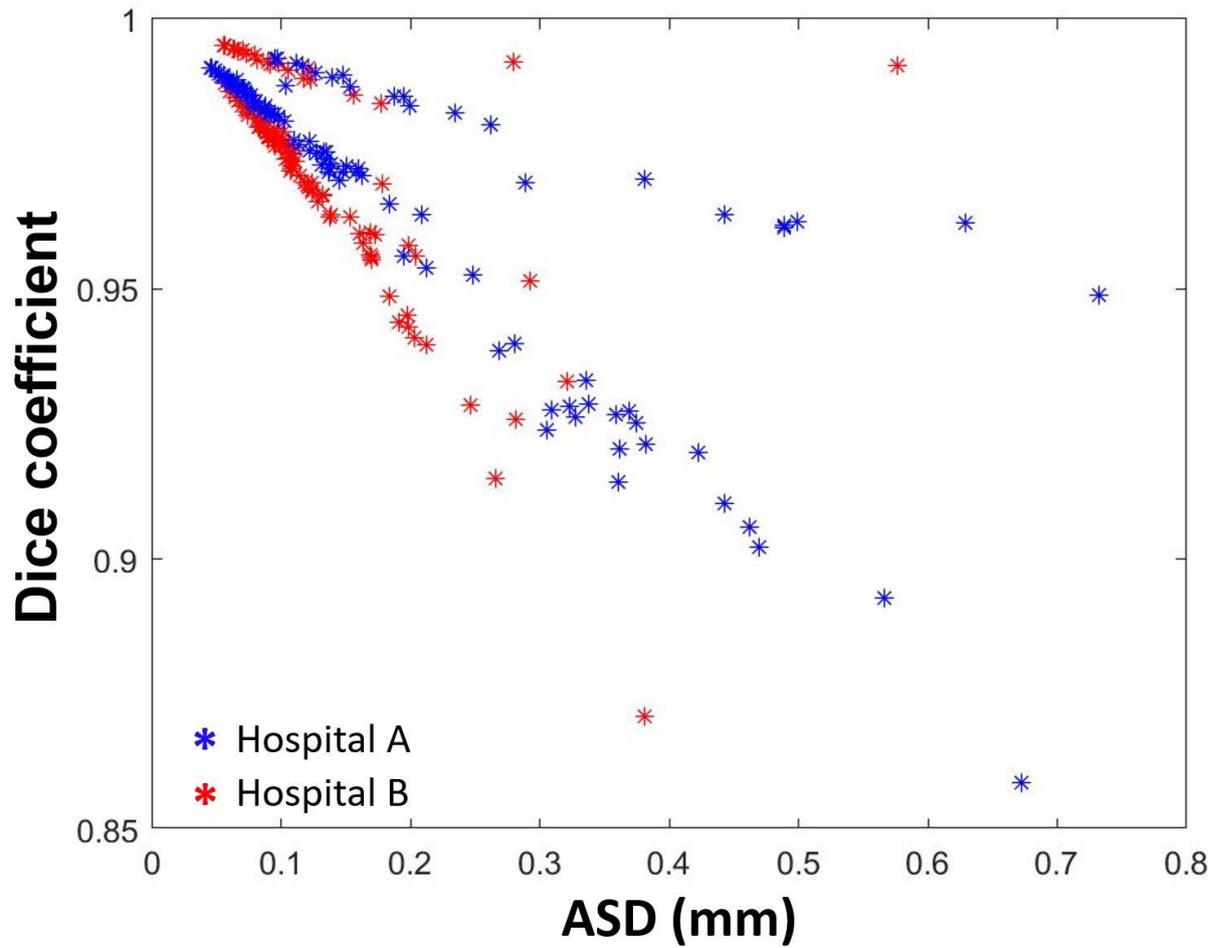

**Fig. 5** Scatter plots of the Dice coefficient and average symmetric surface distance (ASD) for all five regions of the calibration phantom. Results of hospital A and hospital B are indicated in blue and red, respectively

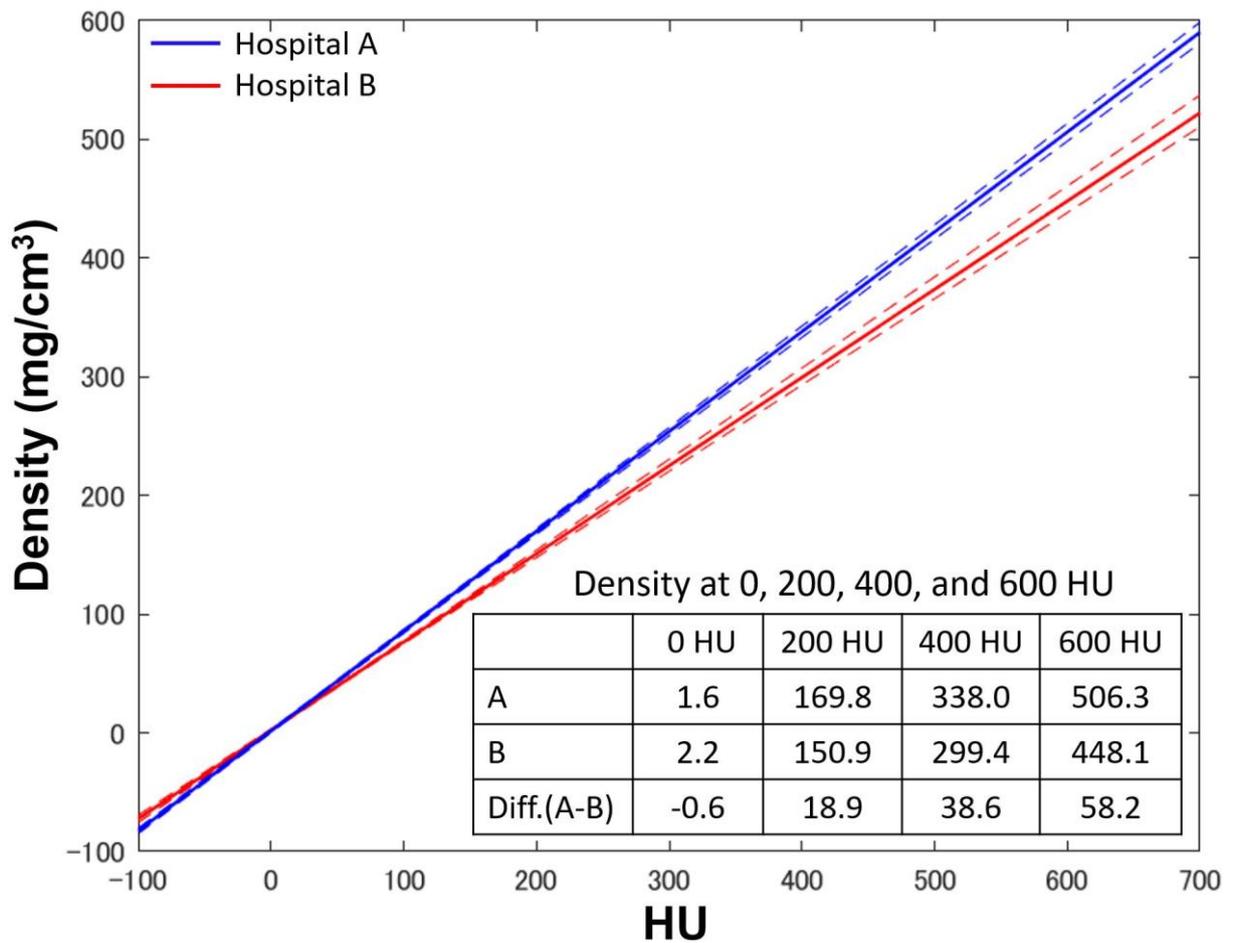

**Fig. 6** Relationship between radiodensity (horizontal axis) and tissue density (vertical axis) in the models for hospitals A (blue) and B (red) from −100 to 700. Solid lines indicate the medians, and dotted lines indicate the interquartile ranges. The table in the lower right corner shows tissue densities calculated for 0 HU, 200 HU, 400 HU, and 600 HU with each hospital's model. Diff.: Difference

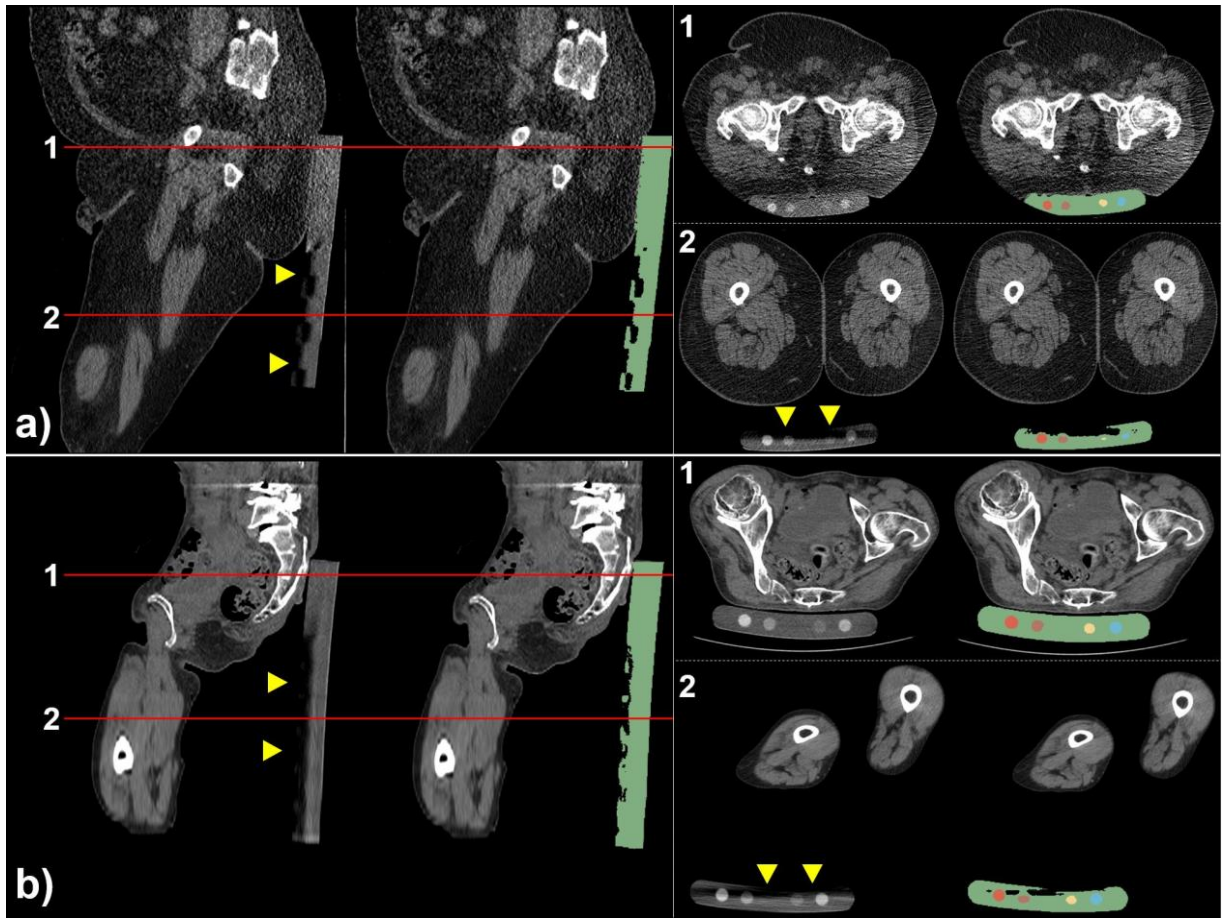

**Fig. 7** CT images and segmentation results of the calibration phantom shown on the sagittal and axial views in two cases that exhibited low correlation coefficients: a) a patient with a bone mass index of 40.6, and the model that had the lowest correlation coefficient among the test cases (0.9863), and b) that of the model that had the second lowest correlation coefficient (0.9908). In the first case, the border of the calibration phantom was difficult to distinguish on the axial slice at the level of the femoral head head (right upper row, red horizontal line labeled 1 on the sagittal view) and with halation (indicated by yellow triangles) was on the axial slice at the center of the femur (right lower row, red horizontal line labeled 2 on the sagittal view), which led to segmentation

errors. In the second case, while the calibration phantom is well segmented on the axial slice at the head level (right upper row), there is halation at the center of the femur, leading to segmentation deficiency